%% file: main.tex
\documentclass[10pt,twocolumn]{article}
\usepackage[letterpaper,margin=0.72in]{geometry}
\usepackage[T1]{fontenc}
\usepackage{mathptmx}          
\usepackage{graphicx}
\usepackage{amsmath}
\usepackage{amssymb}
\usepackage{xcolor}
\usepackage{url}
\usepackage[hidelinks]{hyperref}
\usepackage{caption}
\usepackage{subcaption}
\usepackage{authblk}
\usepackage{booktabs}
\usepackage{multirow}
\usepackage{stfloats}   
\usepackage{placeins}  
\usepackage{titlesec}
\titlespacing*{\section}      {0pt}{1.8ex plus .2ex minus .2ex}{1.1ex plus .1ex}
\titlespacing*{\subsection}   {0pt}{1.5ex plus .2ex minus .2ex}{0.8ex plus .1ex}
\let\OldSubsection\subsection
\renewcommand{\subsection}{\FloatBarrier\OldSubsection}

\title{\bfseries\LARGE Testing Between the Test Cases: Proving
End-to-End Steering in Conditions You Never Drove}

\author[1]{Menuka Ghalan}
\author[1]{Charles Rodgers}
\author[2]{Zachary D. Asher}
\affil[1]{Department of Computer Science, Western Michigan University, Kalamazoo, MI, USA}
\affil[2]{Department of Mechanical and Aerospace Engineering, Western Michigan University, Kalamazoo, MI, USA}
\date{}

\begin{document}
\twocolumn[
\begin{@twocolumnfalse}
\maketitle
\begin{abstract}
\noindent AI-based automated vehicle testing is challenging because a model that passes every test condition can still fail in the real world. Formal verification offers a way to directly address this gap. On a simulated highway and an arterial road we trained two small end-to-end steering networks each in CARLA, one on clear conditions alone and one on clear, fog, night and low sun. All four models were driven against a $2.19$\,ft lane-departure budget. Without driving again, we used bound propagation, a formal method that reads the trained weights, to compute how far steering can drift at every disturbance strength between two captured images. One calculation covers more than a campaign could drive: on the arterial it spans $133$ poses, where ten intensities each would be $10^{133}$ combinations, in minutes on one GPU. Not only did formal verification find conditions that broke the clear-trained policy without simulation testing, it provided some preliminary evidence for potential failures between the test cases. Our overall conclusion is that formal verification is a viable complement to simulation, and could be adopted as a part of verification and validation for automated driving.
\end{abstract}
\vspace{0.8em}
\end{@twocolumnfalse}
]

\section{Introduction}

Autonomous vehicle (AV) development has moved from rule-based pipelines to data-driven ones, where safety and scalability have been the forces behind every shift along the way \cite{liu2025avsurvey}. End-to-End (E2E) neural networks are the current expression of that shift. In an E2E model, a single network maps camera pixels directly to a driving command such as a behavior-cloned steering controller \cite{bojarski2016end} or a reinforcement-learned lane follower \ \cite{kendall2019learning}. The argument for E2E design is that optimizing the whole software stack against the driving objective avoids the errors that compound through separately trained perception, tracking and planning modules and they are easier to scale \cite{chen2024end}. Modern E2E models now extend to jointly optimized stacks that still expose intermediate representations \cite{hu2023planning} and to vectorized planners built for efficiency \cite{jiang2023vad}. The datasets used to train and benchmark these policies have grown accordingly \cite{liu2025tade2e}.

Compliance is where the E2E design becomes awkward. ISO 26262 \cite{iso26262} allocates functional-safety requirements to components whose failure modes can be enumerated and it needs adapting before it can accommodate machine learning components \cite{salay2017iso26262}. ISO 21448 \cite{iso21448} addresses the Safety of the Intended Functionality (SOTIF), the hazards that remain when nothing has malfunctioned, while ISO 8800 \cite{iso8800} and UL 4600 \cite{ul4600} extend the argument to AI components specifically and to autonomous products as a whole. The difficulty common to all of them is that the E2E policies that perform best are the ones least amenable to the enumeration these standards assume.

The prevailing answer today is empirical, a test-and-patch cycle whose evidence operators assemble into an explicit safety case \cite{favaro2023waymo, waymo2020framework}. Simulation carries much of that load, and the Car Learning to Act (CARLA) simulator  is the standard open platform for it \cite{Dosovitskiy17, vivan2021no}. What simulation cannot do is enumerate. Change one parameter and the whole campaign runs again. Data-driven policies generalize poorly on the long-tail scenarios that fall outside their training distribution \cite{liu2025avsurvey}. A current solution is to collect dedicated adverse-condition datasets and utilize them in training \cite{Sakaridis_ACDC_ICCV_2021, Sakaridis_2026_TPAMI}. This has been extended more recently to generated inputs that enlarge the training distribution \cite{nvidia2026omnidreams}. Camera-driven policies also face inputs no campaign samples so including adversarial examples that transfer across architectures is recommended \cite{AdversarialTransferability2026}. Despite these efforts, unanticipated physical conditions still disrupt exactly the decision windows a statistical safety case assumes are covered, because no amount of input data covers the uncountably many ways a disturbance can manifest. The problem is not that the test cases are wrong, but that they are finite.

Formal verification is the alternative. Testing runs a system on chosen inputs and reports what happened on them. Formal verification takes a set of inputs, described mathematically, and computes a guarantee holding for every input in that set, including ones that have not been physically run. For a neural network the practical form is bound propagation: the method reads the trained weights, pushes an input range through the layers, and arrives at a range the output cannot leave. If that range sits inside a safe limit, every input in the set is safe, and the argument is arithmetic.

The gap from testing is not one of degree. A disturbance that varies continuously has uncountably many settings; a testing campaign visits finitely many, and so does a generated one. Older surveys of industrial practice put formal methods mostly at specification and design, with tooling among the barriers \cite{woodcock2009formal} even though model checking is long established for finite-state control logic \cite{clarke2018model}. For neural networks the tooling has matured quickly, from complete satisfiability-modulo-theories (SMT) methods \cite{katz2017reluplex, katz2019marabou} to reachability frameworks for learning-enabled systems \cite{tran2020nnv}, and on to closed-loop controllers by partitioning the workspace so the imaging function is affine on each piece \cite{sun2019nncontrolled}.

What limits formal methods is the disturbance modeling, not the mathematical solvability. Verifiers are overwhelmingly exercised on $\ell_p$ balls around an image \cite{wang2021beta, singh2019abstract}, a set whose radius is chosen for tractability and whose tightest relaxation is itself bounded \cite{salman2019convex}. A ball large enough to contain a night-time image also contains images no camera can form, so a bound over it says nothing about real-world disturbances experienced by automated vehicles. While verification is one thread in a wider effort spanning autoformalization \cite{szegedy2020autoformalization}, interpretability \cite{anthropic2025tracing} and statistical safety-case confidence \cite{bishop2022bootstrapping}, we target the perception-to-control bound, which is directly relevant for automated driving using E2E models.

In this work we apply neural-network formal verification to E2E steering under degraded visibility, and we make three contributions. All of the evidence is from the CARLA simulator, on two roads, with a camera as the only sensor and steering as the only learned command. We show that a bound taken only at the conditions a campaign captures is not sufficient, because it clears a policy whose worst case lies at an intermediate strength. We give a disturbance family that brings those intermediate intensities within a verifier's reach. And we reconcile a per-frame bound with closed-loop driving using vehicle dynamics. We seek neither a new verification algorithm \cite{althoff2007online} nor a new planning benchmark \cite{althoff2017commonroad}, but the mapping from a physical disturbance to a set a neural network verifier can bound. The result is a practical route from a physical disturbance to safety-case evidence about conditions nobody drove, using a method automotive validation has barely begun to adopt where one bound covers more of the operating envelope than a test program can drive.

\section{Related Work}
Formal verification of automated driving models addresses the gap between empirical AI performance and strict safety standards. Koopman and Wagner argue in 2018 that the confidence deployment requires cannot be reached by vehicle-level testing alone \cite{koopman2018framework}, having first mapped where the ISO 26262 V-model strains against autonomous driving in 2016 \cite{koopman2016challenges}. For example, it has been shown that to demonstrate a reduction on an interstate with one fatal accident per $662$ million km at $5\%$ significance would require some $6.62$ billion test kilometers \cite{wachenfeld2016release}. Wagner has since carried that accounting to fleet-scale safety-case assessment \cite{wagner2026agile}. Burgio et al. \cite{burgio2024open} detail the open challenges of verification, in which AVs are heterogeneous systems combining white-box and black-box AI components. Mitra et al.\ \cite{mitra2025formal} show how the high-dimensional nature of visual perception data fundamentally limits traditional mathematical guarantees. Pan et al.\ \cite{pan2025automating} pair large language models with model-driven engineering, deriving formal models from event-chain descriptions to validate generated components at the system level. But formal verification applied to E2E driving policies, against the disturbances a vehicle actually meets, remains largely unexplored.

Foundational formal methods have, by contrast, verified substantial non-AI software. The interactive theorem provers Coq \cite{bertot2004coqart}, Isabelle/HOL \cite{nipkow2002isabelle} and Lean \cite{moura2015lean} carry machine-checked proofs for the CompCert C compiler \cite{leroy2009compcert} and the seL4 microkernel \cite{klein2009sel4}, and they remain relevant for certifying deterministic components under ISO 26262 \cite{iso26262}. They target explicit rule-based logic, so deep learning needs a different paradigm.

When applying formal methods, the disturbance model, not the solver, is what limits applicability to automated driving. Bernardeschi et al.\ \cite{bernardeschi2025verifying} evaluate E2E driving robustness against imperceptible image modifications and report probabilistic guarantees, but macroscopic physical deviations such as global illumination drops remain open. Most directly related, Mohapatra et al.\ \cite{Mohapatra_2020_CVPR} verify robustness against parametric semantic perturbations such as brightness and contrast in place of $\ell_\infty$ pixel balls. We adopt that parametric view, take the endpoints from simulator renders at a fixed camera pose instead of an analytic image model, and then reconcile the per-frame bound with closed-loop driving.

Our verifier is the CROWN family \cite{zhang2018crown, xu2020auto_lirpa} with input-space branch and bound. Bound propagation is white-box in the weights: it needs the trained parameters and the architecture which is a different access requirement from testing and is open to a developer or to any assessor given the weights. Our disturbance family is built from pose-paired simulator renders, so for this work the bound needs a renderer even though it never needs to drive. Semidefinite relaxations such as SDP-CROWN \cite{chiu2025sdpcrown} tighten bounds on high-dimensional $\ell_2$ balls, a regime our one-parameter set does not enter as it stands, though the wider set we propose in the Conclusion would. The approach complements verifiers based on abstract interpretation \cite{gehr2018ai2}, control reachability \cite{ivanov2019verisig, julian2021reachability} and uncertainty-aware analysis of E2E driving \cite{michelmore2020uncertainty}.

\input{sec_methodology}

\input{sec_results}

\section{Conclusion}
We trained two small camera-only steering models in simulation, one on the clear condition and one on clear, fog, night, and low sun conditions. This was done for the Town04 highway and the Town06 arterial. We the apply formal verification using a straight line in image space between two frames captured at the same camera position, so a single number sets how strong the disturbance is and the verifier can be asked about all of it at once. The verifier bounds the steering error that persists along the route which causes failures. It correctly predicted every highway outcome and also identifies potential failure points that were never tested in closed loop.

This paper provides evidence that formal verification viable for a narrow aspect of automated driving verification. Formal verification covers more conditions in a few minutes than a test program could drive in its lifetime, but the family of policies under test must be developed as a mathematical set. Future work involves widening the mathematical set to include more disturbances such as rain, snow, glare, and dust, as well as combinations of these disturbances. Additionally we seek to continue testing the predicted failures from formal verification to further expand its utility.

\section*{Data Availability}
The pipeline, instruments, checkpoints and every artifact behind the numbers in this paper are released as a versioned software record whose concept DOI resolves to the latest version \cite{ghalan_steering_code}, mirrored on GitHub under the AD-Assurance-Lab organization. The captured frames the bounds are computed on are published separately \cite{steering_captures_dataset}.

\section*{Abbreviations}
{\footnotesize
\begin{tabular}{@{}ll@{}}
AI     & artificial intelligence \\
AV     & autonomous vehicle \\
CARLA  & Car Learning to Act \\
CNN    & convolutional neural network \\
CTE    & cross-track error \\
DAgger & dataset aggregation \\
E2E    & end-to-end \\
FC     & fully connected \\
FOV    & field of view \\
GPU    & graphics processing unit \\
MSE    & mean squared error \\
ODD    & operational design domain \\
PI     & proportional-integral \\
PPC    & pure-pursuit controller \\
ReLU   & rectified linear unit \\
RGB    & red, green, blue \\
SDP    & semidefinite programming \\
SMT    & satisfiability modulo theories \\
SOTIF  & safety of the intended functionality \\
\end{tabular}}

\vspace{0.6em}

\makeatletter \let\oldbibliography\thebibliography
\renewcommand{\thebibliography}[1]{%
\oldbibliography{#1}%
  \setlength{\itemsep}{0pt}%
  \setlength{\parsep}{0pt}%
  \setlength{\parskip}{0pt}%
  \setlength{\labelsep}{0.3em}%
  \setlength{\labelwidth}{1.6em}%
  \setlength{\leftmargin}{1.9em}%
} \makeatother {\footnotesize \bibliographystyle{IEEEtran}
\bibliography{references}
}

\end{document}

%% file: sec_methodology.tex
\section{Methodology}
The goal is a per-frame formal verification result whose verdict matches what the vehicle does in closed-loop simulation, without actually running the simulation. This is applied in two operational design domains (ODDs): a highway and an urban arterial in CARLA (\S\ref{sec:twostudies}). The policy is a small distilled student, since bound-propagation cost scales with size (\S\ref{sec:policies}). Disturbances come from rendered frames, not an analytic model, and the verified set is the one-parameter family between two of them (\S\ref{sec:disturb}). The safety test uses the sustained steering deviation rather than the peak, which is what a lane-keeping limit is derived for (\S\ref{sec:tol}).

\subsection{Simulator, Vehicle, and ODD}
\label{sec:twostudies}\label{sec:determinism}
The study runs in the open source CARLA simulator version 0.9.16 \cite{Dosovitskiy17}. A simulator is required here, because the disturbance family of \S\ref{sec:disturb} needs a clear frame and a degraded frame from identical camera poses, which a real drive cannot supply. The ego vehicle is a Tesla Model~3 from CARLA's standard blueprint library, a passenger car whose bounding box sets the lane-departure budget of \S\ref{sec:tol}. To keep the study simple, the network steers and nothing else: longitudinal speed is held at a constant $v_x = 8.9408$\,m/s ($20$\,mph) by a proportional-integral (PI) controller at a $5$\,Hz control rate ($\Delta t = 0.2$\,s), and the only sensor is a front RGB camera at $640\times480$ with a $90^\circ$ field of view. Repeatability is not free in CARLA, which is built on a game engine tuned for smooth graphics rather than for reproducing a lap. We have thoroughly addressed this by ensuring the simulator is stepping on a fixed timestep, each steering command lands on the tick it was issued for, and background streaming of texture detail is switched off.

\begin{figure}[!hb]
  \centering
  \includegraphics[width=\columnwidth]{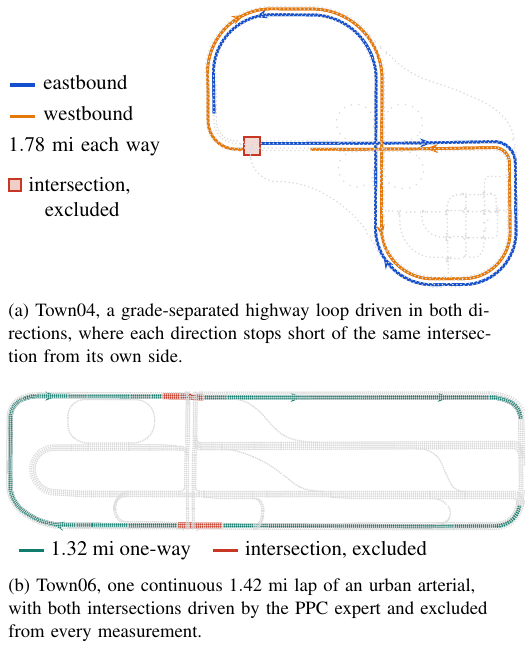}
  \caption{The two roads: a grade-separated highway loop driven both ways, and one
  continuous lap of an urban arterial whose two intersections are excluded from every
  measurement.}
  \label{fig:routes}
\end{figure}

We chose two CARLA maps that are different ODDs in the vocabulary of ISO~34503 \cite{iso34503}, where road type and road geometry are first-class ODD attributes. Town04 is a grade-separated highway loop and Town06's outer loop is an urban arterial with tighter corners. In the terms of Koopman and Fratrik \cite{koopman2019safeai} these are two separate ODDs and their results are likely to disagree. Both routes are drawn in Fig.~\ref{fig:routes}.

The Town04 highway study was completed first, and it is where the criterion and its one calibrated constant were chosen: the closed-loop reaction horizon $T_{\text{cl}}$. It stands for how long a steering error is assumed to persist before the vehicle either corrects it or leaves its lane because of it. The Town06 arterial study then used that same value, without refitting it. A lap is one traversal of the full route, and it fails if the vehicle leaves its lane at any point along the way. A test case is one policy driven under one condition. Each test case is driven three independent laps, and those three laps are certified together as a single unit. The three laps helped us diagnose many internal errors as we developed the test case procedure.

\subsection{AI Model Development}
\label{sec:policies}
\begin{figure*}[!t]
  \centering
  \includegraphics[width=\textwidth]{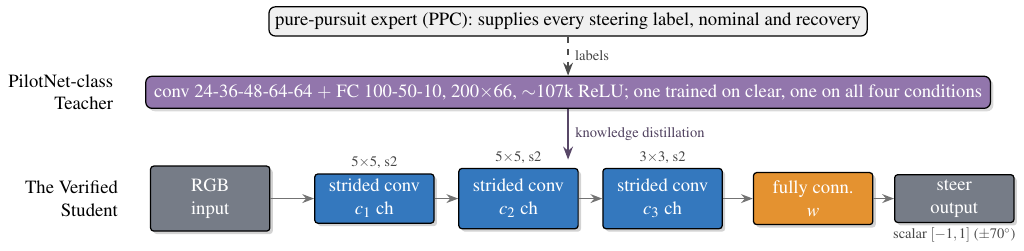}
  \caption{All four models share this topology, differing only in the widths tabulated in
  Table~\ref{tab:models}. Each is distilled from a PilotNet-class teacher trained on
  pure-pursuit labels, and only the student model is verified.}
  \label{fig:netarch}
\end{figure*}

There is a well-known tension in formally verifying an E2E lane-keeping policy: bound-propagation cost scales with network size, so a verifiable policy must be small, yet a small network is hard to train to expert quality. We resolve this tension with a teacher$\rightarrow$student design (Fig.~\ref{fig:netarch}) in which the large teacher is only a distillation source and is never verified. A PilotNet-class network \cite{bojarski2016end} of 5 convolutional $+$ 4 fully-connected layers, rectified-linear (ReLU) units only, on a $200\times66$ red-green-blue (RGB) crop ($\sim$107k ReLU neurons) is used for the teacher.

The student is a ReLU-only convolutional neural network (CNN), 3 strided convolutions and 2 fully-connected (FC) layers, whose region of interest is fixed from ground-truth segmentation. By cropping tightly, we keep lane markings legible at low resolution and also enable a small RGB input size. Table~\ref{tab:models} provides the size details of each of the four models, ranging from $5{,}152$ ReLU neurons on Town04 to $101{,}888$ on Town06, which needed a larger input to pass closed-loop testing. Each road utilizes two students distilled from its own teacher, one trained on the clear condition alone and one on the four conditions of \S\ref{sec:disturb}, clear, fog, night and low sun. We call the final models the clear-trained and the mixed-trained model throughout. The mixed-trained model is wider on both roads because it represents four conditions rather than one. Network width is more effective than depth, which at a matched neuron count widens the certified bounds by $2.3$ to $3.7\times$.

\begin{table}[!hb]
\centering
\caption{The four verified models. All share the topology of Fig.~\ref{fig:netarch} and
differ only in these numbers.}
\label{tab:models}
\footnotesize\setlength{\tabcolsep}{3.5pt}
\begin{tabular}{llcccr}
\toprule
Road & Student & Input & $(c_1,c_2,c_3)$ & FC $w$ & ReLU \\
\midrule
\multirow{2}{*}{Town04} & clear-trained & $84{\times}28$  & $(8,16,16)$  & $32$  & $5{,}152$ \\
                        & mixed-trained & $84{\times}28$  & $(24,48,48)$ & $96$  & $15{,}456$ \\
\midrule
\multirow{2}{*}{Town06} & clear-trained & $168{\times}56$ & $(16,32,32)$ & $64$  & $50{,}944$ \\
                        & mixed-trained & $168{\times}56$ & $(32,64,64)$ & $128$ & $101{,}888$ \\
\bottomrule
\end{tabular}
\end{table}

To begin model development, the teacher is first fitted by behavior cloning on PPC-derived expert camera and label pairs over the route, then refined by Dataset Aggregation (DAgger) \cite{ross2011dagger}. Through the DAgger process, the current model iteration drives the full route and when it wanders off the lane, the PPC is queried for the correct recovery action which is then added to the training set. The teacher is retrained on the aggregate and the process repeats until the teacher drives within budget. The student is then distilled from the converged teacher, and runs its own DAgger rounds until it too drives within budget.

We learned several lessons throughout this process. The mixed model DAgger rounds need a warm start, meaning that each round fine-tunes the checkpoint from the round before it, at a reduced learning rate, instead of training again from scratch. Additionally, when a policy needs more capacity, it is better to widen it than to feed it a larger image, because widening adds parameters without enlarging the set the verifier has to bound. Lastly, it pays to keep refining the teacher through DAgger after it already drives within budget: the student distilled from it comes out $2.7\times$ closer.

\subsection{Disturbance Modeling}
\label{sec:disturb}
Each disturbance must be a set the verifier can bound, and it must provoke the same response from the policy as the real condition does. We therefore take the clear and disturbed images from the exact same point in the simulator itself rather than from a photometric model or a real-world dataset. For each pose $p$ on the route we capture a clear frame $x_p^{\text{clear}}$ and a condition frame $x_p^{\text{cond}}$. Each one is the array of pixel values the network receives on a $0$ to $1$ brightness scale rendered at full sensor resolution and only then cropped and downsampled. These captured frames are the ones the verifier reads, and we check per pose that they are what the model saw while driving.

\begin{figure}[!hb]
  \centering
  \includegraphics[width=\columnwidth]{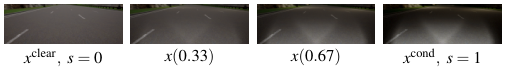}
  \caption{Eq.~\eqref{eq:family} as pictures. The two ends are renders at the same camera
  pose, the two frames between them are blends, and $s$ dials the strength.}
  \label{fig:family}
\end{figure}

Between those two endpoints we declare
\begin{equation}
\label{eq:family}
    x_p(s) = x_p^{\text{clear}} + s\,\bigl(x_p^{\text{cond}} - x_p^{\text{clear}}\bigr),
    \qquad s \in [0,1],
\end{equation}
in which $s$ is a single dial for the strength of the disturbance, so Eq.~\eqref{eq:family} stands for a whole family of disturbances rather than one. At $s=0$ the image is the clear frame, at $s=1$ the rendered disturbance is at full strength, and in between the disturbance is a blend of the two. Figure~\ref{fig:family} shows how this construction captures an entire family of disturbances through one equation. The verifier is then asked about one unknown number instead of thousands of independent pixels, which is what makes it possible to cover every strength in the range at once instead of the single strength a driven lap happens to sample. Cropping and downsampling are linear, so blending the two small network inputs gives exactly the image we would get by blending the two full-resolution renders and shrinking the result. The dial $s$ stays inside $[0,1]$ because outside it the pixel values run past black or white.

\begin{figure}[!hb]
  \centering
  \includegraphics[width=\columnwidth]{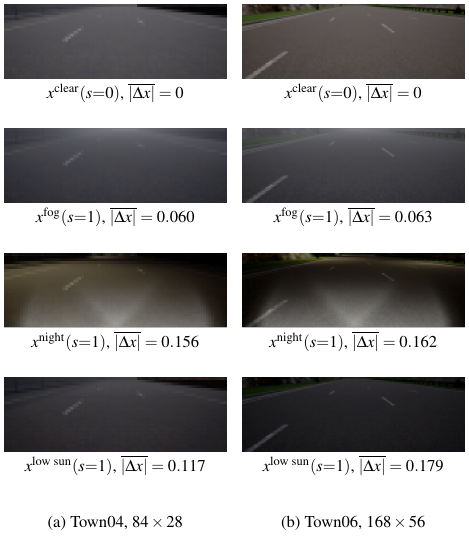}
  \caption{The four conditions as each network receives them. Each panel gives
  $\overline{|\Delta x|}$, the mean absolute change from the clear condition, averaged over every
  pixel and every pose on the $0$ to $1$ brightness scale.}
  \label{fig:conditions}
\end{figure}

We study four conditions on this axis: clear (the $s=0$ baseline), fog, night, and low sun. Figure~\ref{fig:conditions} shows each as the network receives it, with the mean absolute change it makes to a pixel. Low sun sounds the mildest and is the strongest disturbance on the arterial, because that sun angle puts the whole road in terrain shadow with no artificial light. With the sun below the horizon the simulator switches the vehicle's headlights on and lights the road directly, brightening about a quarter of the pixels while the rest go darker still. The whole road being in shadow is also why we call the condition low sun rather than shadows.

Three ideas we tried and abandoned are worth noting. The first was an analytic Koschmieder fog model \cite{Narasimhan2002Vision} which is the attenuation-plus-airlight form that single-image dehazing inverts \cite{He2010DarkChannel}. Its appeal is that the one free parameter is physical. It reproduced CARLA's fog images respectably, at a road-ROI $R^2 = 0.848$, and was still unusable, because looking like real fog is not the same as steering the policy like real fog. The network reacted far more strongly to the model than to the rendered fog it stood in for, and analytic models of night \cite{wei2018deep} fail that test the same way. The second was a real-world dataset. The Adverse Conditions Dataset with Correspondences (ACDC) \cite{Sakaridis_ACDC_ICCV_2021} pairs every adverse frame with a matching view of the same scene in clear conditions, but the two come from separate drives, so the difference between them carries viewpoint and scene change as well as the condition, where Eq.~\eqref{eq:family} needs it to be the condition alone. The third was rain, which CARLA renders with a temporally random element that a two-endpoint family cannot represent.

\subsection{The Safety Criterion}
\label{sec:tol}
Success for a lane-keeping policy means no part of the vehicle leaves its lane. With the vehicle centered, the room on either side is half the lane width minus half the vehicle width. The cross-track error (CTE), measured from vehicle center to lane center, therefore has a budget of
\begin{equation}
\label{eq:budget}
    \text{CTE}_{\text{budget}} = \frac{w_{\text{lane}} - w_{\text{veh}}}{2} .
\end{equation}
The lane width $w_{\text{lane}} = 3.500$\,m is measured constant along both routes, and the vehicle width $w_{\text{veh}} = 2.164$\,m is the CARLA bounding box of the Tesla Model~3, mirrors included. That leaves $0.668$\,m ($2.19$\,ft) of allowable CTE.

Converting that distance into a steering limit starts from the kinematic bicycle model. The vehicle holds the constant longitudinal speed $v_x$ and a steering angle $\delta$ on a wheelbase $L$ which turns it at the yaw rate $\dot\psi$. This relationship is then linearized for small steering angles as
\begin{equation}
\label{eq:bike}
    \dot\psi = \frac{v_x}{L}\tan\delta \approx \frac{v_x}{L}\,\delta .
\end{equation}
Here $v_x = 8.9408$\,m/s (discussed in \S\ref{sec:twostudies}) and the Tesla Model~3's wheelbase is $L = 3.005$\,m. Integrating Eq.~\eqref{eq:bike} with a constant steering error $\Delta\delta$ in place of $\delta$ gives the heading error that steering error produces after $t$ seconds,
\begin{equation}
\label{eq:heading}
    \Delta\psi(t) = \int_0^t \frac{v_x}{L}\,\Delta\delta\,d\tau = \frac{v_x}{L}\,\Delta\delta\,t ,
\end{equation}
where $\tau$ is the integration variable. The speed itself does not change. The heading error only tilts the velocity off the lane direction, so the component across the lane is $\dot y = v_x \sin\Delta\psi \approx v_x\,\Delta\psi$. Integrating that lateral velocity, with Eq.~\eqref{eq:heading} substituted for $\Delta\psi$, gives the lateral offset the steering error accumulates,
\begin{equation}
\label{eq:tol}
    y(T) = \int_0^T \dot y\,dt
         = \int_0^T \frac{v_x^2}{L}\,\Delta\delta\,t\,dt
         = \frac{v_x^2T^2}{2L}\,\Delta\delta ,
\end{equation}
where $T$ is how long the steering error persists. Setting $y(T)$ equal to the CTE budget and solving for $\Delta\delta$ gives the largest steering deviation the vehicle can absorb, written the way the network writes steering, as a fraction of its configured limit $\delta_{\max} = 70^\circ$:
\begin{equation}
\label{eq:deltatol}
    \delta_{\text{tol}} = \frac{2L\,\text{CTE}_{\text{budget}}}
                               {v_x^{2}\,T_{\text{cl}}^{2}\,\delta_{\max}} = 0.0120 .
\end{equation}
The value we use for $T$ is $T_{\text{cl}}$, the closed-loop reaction horizon. Fundamentally it represents how long a systematic steering error survives in the frames the model's own driving produces. We fit it once on the Town04 highway, from the steering bias at which laps actually began to leave the lane, and reused that value, $T_{\text{cl}} = 1.85$\,s, unchanged on the Town06 arterial road. The verdicts do not turn on its exact value. Every highway verdict holds across a factor-of-$1.7$ window in $T_{\text{cl}}$ that contains the roughly $1.5$\,s perception-reaction time the driver-response literature reports for surprise events \cite{green2000stop}.

\subsection{Formal Verification}

%
\begin{figure*}[!b]
  \centering
  \includegraphics[width=\textwidth]{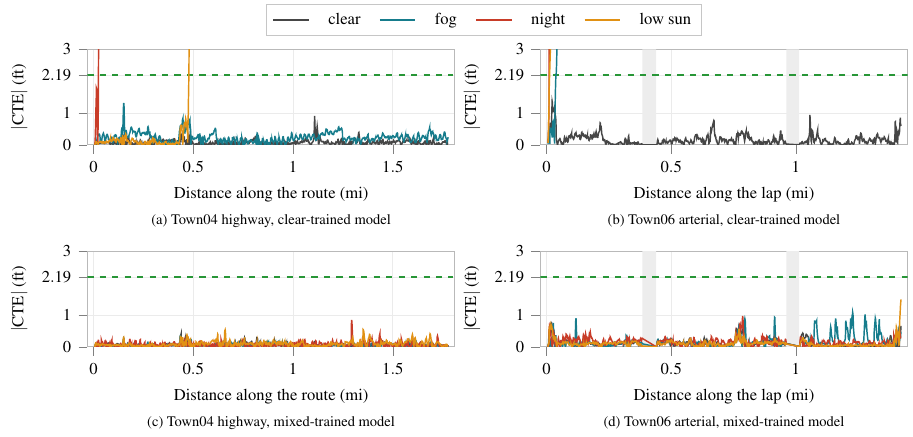}
  \caption{Cross-track error along the route against the
  $2.19$\,ft budget (dashed). Each trace stops the first time it passes $3$\,ft, and the
  gray bands are Town06's two bridged intersections, which no policy steers.}
  \label{fig:cte}
\end{figure*}

Formal verification provides a maximum bound on a property over a declared set. Such a bound answers not what the network outputs on one image, but the most it could output on any image in a set. Hand it the whole family of Eq.~\eqref{eq:family} rather than one disturbance strength at a time, and it returns an interval guaranteed to contain every steering value the network can produce anywhere in that family. This is computed by propagating the set through the layers as algebra rather than as sampled points, with each ReLU replaced by a linear upper and lower envelope. Note that the formal verification literature uses the word `certified' to mean that the bound stayed under a desired threshold. The words `certified' and performance `certificate' are not meant in the regulatory sense because they do not approve a vehicle for road use. Formal verification's evidence can be one part of a safety case, but it is not a substitute for one.

CROWN \cite{zhang2018crown, xu2020auto_lirpa} is the established algorithm we use for formal verification, in its base form rather than the refinements $\alpha$-CROWN \cite{xu2021fast}, $\beta$-CROWN \cite{wang2021beta} and SDP-CROWN \cite{chiu2025sdpcrown}, whose gains come from high-dimensional input sets and from branching over ReLU activations, neither of which a one-parameter family offers. What matters for validation is that the interval it returns is one-sided, meaning that it can be wider than the true range of outputs but never narrower. That looseness grows with the width of the set being bounded, so we do not ask for one bound over the whole family at once. We cut the strength range into sub-intervals, bound each separately, and take the worst which is still a bound on the whole family. Narrower pieces give tighter bounds, which is what makes the result sharp enough to certify a policy that drives cleanly. The number of pieces is a free parameter and we use 16 sub-intervals per pose on both roads.

\label{sec:criterion}
The bound and the budget now meet. The verifier runs at $P$ poses along the whole lap, one every eighth control step, and at each pose it reports how far the disturbance can move the steering away from the value the same network produces on the clear frame as
\begin{equation}
\label{eq:dev}
    \Delta_p(s) = \delta_p(s) - \delta_p(0) ,
\end{equation}
where $\delta_p(s)$ is the steering on the blended image $x_p(s)$ at any strength from the clear frame at $s=0$ to the full condition at $s=1$. The clear frame is a reference and not ground truth. Subtracting it separates what the disturbance does from how well the policy steers in the first place, which the clear-condition drives answer on their own. Equation~\eqref{eq:deltatol} is likewise not a limit on one frame. It is the steering error that would use up the whole lane budget if it were held for $T_{\text{cl}}$ seconds, so what we compare against it has to be a sustained error rather than a spike. We therefore average the per-pose deviations along the lap,
\begin{equation}
\label{eq:mean}
    \bar\Delta(\mathbf{s}) = \frac{1}{P}\sum_{p=1}^{P} \Delta_p(s_p) ,
\end{equation}
which is the steady bias the disturbance leaves in the steering. A spike that reverses sign within a few frames averages away, just as it washes out of the vehicle's path, while a deviation that keeps its sign survives both. Each pose is free to sit at a different strength, so the vector $\mathbf{s} = (s_1,\dots,s_P)$ holds one strength per pose, and the policy is certified when no assignment of strengths pushes the average outside the corridor as
\begin{equation}
\label{eq:crit}
    \max_{\mathbf{s}\in[0,1]^{P}} \bigl|\bar\Delta(\mathbf{s})\bigr| \le \delta_{\text{tol}} .
\end{equation}
CARLA renders each condition uniformly along the road, so this set is deliberately larger than anything a driving campaign can test.

There are a few things to note about this approach. Cutting the strength range into more pieces buys more than a more elaborate bounding algorithm does. The $\alpha$-optimized refinement ran far longer and gave back almost nothing on a family this narrow. How many pieces matters, though. With only four, one policy-condition test case's bound came out at $1.07$ times the tolerance. We also began with the wrong statistic. Testing the single worst frame instead of the road average threw out policies that drive cleanly, and it ranked two test cases in the opposite order. No choice of threshold repairs that.

%% file: sec_results.tex
\section{Results}
The first step in collecting results is to drive every policy until its trained behavior is established in closed-loop simulation (\S\ref{sec:driving}). Additional closed-loop test case verdicts are then compared to formal verification where it was determined that formal verification can correctly predict the outcomes (\S\ref{sec:verif}). Additional analysis of the formal verification shows that it may also be identifying failures between test cases (\S\ref{sec:witness}). We also discuss the ODD difference between the Town04 highway route and the Town06 urban arterial route (\S\ref{sec:odd}).

\subsection{Closed-Loop Driving}
\label{sec:driving}

We proceed to formal verification only once a model holds its lane on every condition it was trained on. The mixed-trained model must drive clear, fog, night and low sun within the CTE budget in closed loop; the clear-trained model must drive the clear condition, though it does not need to be tested in fog, night or low sun before proceeding to formal verification.

Figure~\ref{fig:cte} shows where along each route the CTE accumulates at every control step. A trace ends the first time it passes $3$\,ft, because past that the test case has failed and how much further it goes does not provide any usable information for this work. We observe several interesting behaviors. First the clear-trained model has near zero CTE during clear conditions in Town04 (highway) and Town06 (arterial). The clear-trained model also cannot handle mixed conditions (fog, night, low sun). Likewise the mixed-trained models have near zero CTE during all mixed conditions. However, Town06 does somehow seem to be more challenging because the clear-only model fails faster in mixed conditions and the mixed model has slightly higher CTE overall in mixed conditions.

\subsection{Verification}
\label{sec:verif}

After the closed-loop driving of \S\ref{sec:driving} is complete we are ready to run formal verification. Every model-condition-town test was bounded with CROWN over the whole disturbance family and scored against the criterion of Eq.~\eqref{eq:crit} which is a mathematical process only and involves no further simulation. We are mathematically testing between the test cases. The CROWN verifier algorithm reads the trained weights and the frames already captured. The result is a bound on the worst sustained steering deviation any disturbance in the family can produce.

Table~\ref{tab:cells} puts that formal verification result beside what the same model did in closed-loop testing. The two clear test cases carry a certificate that is vacuous because clear is the point every disturbance is measured from and there is nothing there for the bound to compare against. In Town04 formal verification correctly predicts that the clear-trained model is suitable for fog which is confirmed through closed loop testing. Every other Town04 comparison agrees. Town06 is much more challenging. The worst CTE values are higher overall and the clear-trained model does not work in fog which formal verification also correctly predicts. The interesting case is the mixed-trained model in Town06, which passes closed-loop testing in fog and at night and which formal verification will not certify. Next we can explore this in more detail.

\begin{table}[!hb]
\centering
\caption{Every test case of both studies, one model on one road under one condition. Note that exceeds means the model left its lane.}
\label{tab:cells}
\footnotesize\setlength{\tabcolsep}{4pt}
\begin{tabular}{llccc}
\toprule
        &           & Closed-loop & Worst CTE & Formal \\
Model   & Condition & driving     & (ft)      & verification \\
\midrule
\multicolumn{5}{c}{\emph{Town04, the highway}}\\
\midrule
\multirow{4}{*}{Clear-trained}
 & clear   & \textsc{pass} & $0.91$  & certified$^\dagger$ \\
 & fog     & \textsc{pass} & $1.32$  & certified \\
 & night   & \textsc{fail} & exceeds & not certified \\
 & low sun & \textsc{fail} & exceeds & not certified \\
\cmidrule(lr){1-5}
\multirow{4}{*}{Mixed-trained}
 & clear   & \textsc{pass} & $0.39$  & certified$^\dagger$ \\
 & fog     & \textsc{pass} & $0.37$  & certified \\
 & night   & \textsc{pass} & $0.85$  & certified \\
 & low sun & \textsc{pass} & $0.54$  & certified \\
\midrule[\heavyrulewidth]
\multicolumn{5}{c}{\emph{Town06, the arterial}}\\
\midrule
\multirow{4}{*}{Clear-trained}
 & clear   & \textsc{pass} & $1.37$  & certified$^\dagger$ \\
 & fog     & \textsc{fail} & exceeds & not certified \\
 & night   & \textsc{fail} & exceeds & not certified \\
 & low sun & \textsc{fail} & exceeds & not certified \\
\cmidrule(lr){1-5}
\multirow{4}{*}{Mixed-trained}
 & clear   & \textsc{pass} & $1.30$  & certified$^\dagger$ \\
 & fog     & \textsc{pass} & $1.95$  & not certified \\
 & night   & \textsc{pass} & $1.00$  & not certified \\
 & low sun & \textsc{pass} & $2.19$  & certified \\
\bottomrule
\addlinespace[2pt]
\multicolumn{5}{p{0.97\columnwidth}}{\scriptsize
$^\dagger$Vacuous. Clear is the point every disturbance is measured from, so there is
nothing here for the bound to compare against and the deviation is zero by construction.
The certificate is true but is not informative.}
\end{tabular}
\end{table}

\begin{figure}[!hb]
  \centering
  \includegraphics[width=\columnwidth]{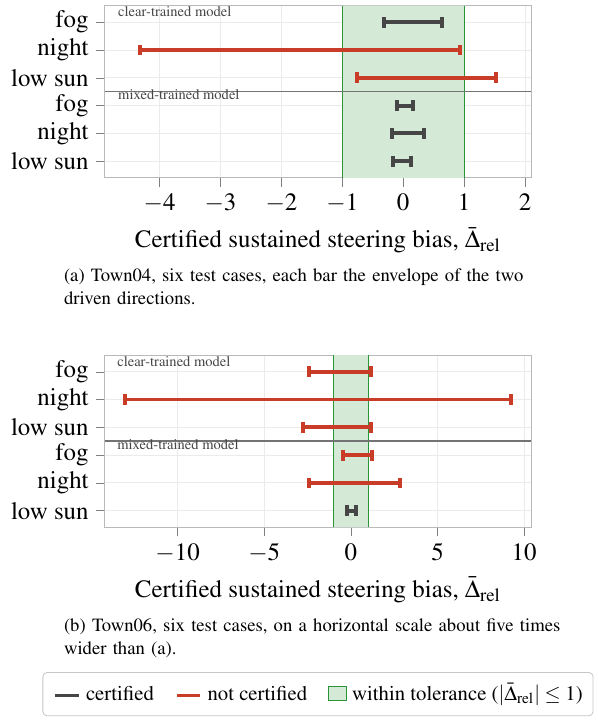}
  \caption{Certified sustained steering bias for every test case in the relative units of
  Eq.~\eqref{eq:rel}, black where the test case certifies and red where it does not.}
  \label{fig:cert}
\end{figure}

Bounds are intuitively reported as road-averaged bias as Eq.~\eqref{eq:mean} divided by the tolerance of Eq.~\eqref{eq:deltatol},
\begin{equation}
\label{eq:rel}
    \bar\Delta_{\text{rel}}(\mathbf{s}) = \frac{\bar\Delta(\mathbf{s})}{\delta_{\text{tol}}} ,
\end{equation}
which means that if a test case has a maximum output between $-1$ and $+1$ it is within tolerance. But a value of $2$ means twice the tolerance (i.e., violates the threshold by $2\times$). Raw steering units would put every number near $0.01$ and would not be comparable between the two roads. The intervals behind Table~\ref{tab:cells} are plotted in Fig.~\ref{fig:cert}, one bar per test case. Certified test cases sit well inside the corridor, refused highway test cases escape it by up to $4\times$, and the arterial's clear-trained night test case escapes by $13\times$. Both clear-trained models sit a long way outside the corridor at night, which matches what they do on the road, where both leave their lane on every run.

Note that each arterial bar in Fig.~\ref{fig:cert} is a statement about $[0,1]^{P}$ with $P=133$, so ten intensities per pose would be $10^{133}$ combinations, which is beyond expensive to drive; it is impossible to drive and impossible to sample. A bar costs a few minutes of GPU time, which is roughly what driving one test case in simulation costs. One for one there is no speedup, but per unit of evidence there is no contest regarding the information that formal verification provides versus closed-loop testing. The bound is a statement about the whole disturbance family, and the closed-loop test is a statement about one disturbance value.

\subsection{Between the Test Cases}
\label{sec:witness}
\label{sec:disagree}
\begin{figure}[!hb]
  \centering
  \includegraphics[width=\columnwidth]{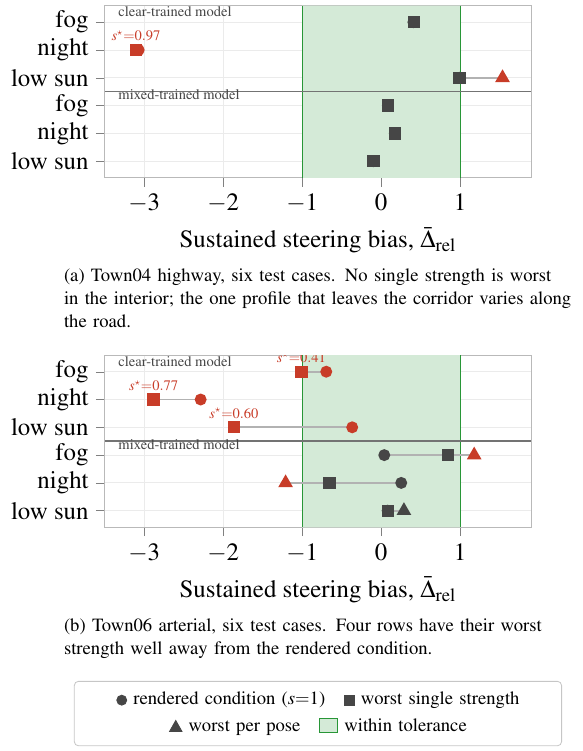}
  \caption{Where inside each disturbance family the worst sustained bias falls, red where
  it leaves the tolerance corridor.}
  \label{fig:witness}
\end{figure}

We can uncover details beyond Fig.~\ref{fig:cert} by going inside the disturbance family and asking specifically what disturbance level (the value $s$) causes the worst output for uncertified test cases. Figure~\ref{fig:witness} adds this information with three markers on each row. The circle is the steering bias at the rendered condition, $s=1$, which is what endpoint testing checks. The square is the worst output from any one strength $s$ of Eq.~\eqref{eq:family} applied to the whole route, marked with the $s^\star$ that produced it, and it is the strongest disturbance a single drive could apply. The triangle is the worst the road average can reach when the strength is free to change from one pose to the next, every pose pushing the steering the same way so nothing cancels. It is larger than anything a single drive can apply, since no disturbance rendering setting varies along a road pose by pose. A triangle is drawn only where the single-strength search left a test case unresolved, which is why panel (a) has just one, on low sun. Panel (a) is the Town04 highway, and its circles and squares coincide in all six test cases: no single strength anywhere in the family is worse than the rendered condition, so driving the condition would have found every failure a single strength can produce. The one low-sun triangle is the exception, and it lies outside the corridor.

Panel (b) is the arterial, and the markers come apart. Let's start with the clear-trained model under fog and under low sun. At the rendered condition the bias sits comfortably inside the corridor, at $0.69$ and $0.37$, which are the two circles, so a bound computed only there would have issued two clean certificates. Both test cases leave the lane on every attempted lap and the squares explain it. The worst single strength is not the rendered one but an intermediate one, $s^\star=0.41$ under fog and $s^\star=0.60$ under low sun, where the bias reaches $1.01$ and $1.87$ and leaves the corridor. The failure was between the test cases the whole time, and quantifying over the family is what found it. Now the two test cases that seemed to contradict the method, fog and night on the mixed-trained model. Their circles and squares are both inside the corridor, which is why the model drives clean: a lap applies one strength to the whole road, and no single strength breaks it. Their triangles are outside, at $1.179$ and $-1.210$. Those are not artifacts of a loose bound but exhibited profiles, strengths assigned pose by pose at which the model's own steering leaves the corridor, so no sound method could certify either test case. The bound and the drive are answering different questions, and the bound is asking the larger one.

The details regarding the AI model failures at the intermediate disturbance levels are worthy of their own closed-loop testing campaign to further expand and improve the application of formal verification to the problem of autonomous driving. As this research continues to evolve, this will be a focus in future studies. Preliminary results suggest that formal verification method is able to identify failures that are not captured by the closed-loop testing, and this information could be used to improve the robustness of the AI models directly.

\subsection{The Difference Is the ODD}
\label{sec:odd}
We explored several alternatives to explain the differences in Town04 and Town06 results. It is not the network size, because the fog bounds get wider as the network shrinks rather than narrower. It is not the input projection either, since capturing the arterial road at the highway's own resolution still leaves fog far outside the tolerance. And it is not one unlucky draw: a pre-registered sweep of sixteen independently distilled mixed-trained models, two widths and eight seeds, drove $83$ laps against a stricter gate than the ledger's and none passed, with fog rejecting every seed that reached it. The teacher in Town06 is not comfortable in fog either, holding $1.53$--$1.68$\,ft against the $2.19$\,ft budget, at $70$--$77\%$ of it. So although fog is drivable on this road, the models struggle with it more than capacity or the random draw explains.

What is left as a variable is the road itself. Road type and geometry are first-class ODD attributes under ISO~34503, and the two routes differ in exactly the attributes \S\ref{sec:twostudies} chose them to differ in: the arterial is $74$--$79\%$ straight against the highway's $51$--$56\%$, with a minimum radius of $22$--$27$\,m against $45$--$63$\,m. Nothing carries from one combination of road attributes to another for free, so the arterial is a domain adjacent to the highway rather than a replication of it, and fog on it is harder to certify at every capacity and resolution we measured.